\documentclass[conference]{IEEEtran}
\usepackage{amsmath,amsfonts}
\usepackage{algorithmic}
\usepackage{algorithm}
\usepackage{array}
\usepackage{makecell}
\usepackage[caption=false]{subfig}
\usepackage{textcomp}
\usepackage{stfloats}
\usepackage{url}
\usepackage{verbatim}
\usepackage{graphicx}
\usepackage{cite}
\usepackage[hidelinks]{hyperref}

\makeatletter
\newcommand{\linebreakand}{%
  \end{@IEEEauthorhalign}
  \hfill\mbox{}\par
  \mbox{}\hfill\begin{@IEEEauthorhalign}
}
\makeatother

\begin{document}

\title{Natural Language Processing with Commonsense Knowledge: A Survey}

\author{\IEEEauthorblockN{Yubo Xie}
\IEEEauthorblockA{\textit{College of Information Engineering}\\
\textit{Shanghai Maritime University}\\
yuboxie@hotmail.com}
\and
\IEEEauthorblockN{Zonghui Liu}
\IEEEauthorblockA{\textit{Independent Researcher} \\
zonghui.liu@hotmail.com}
\and
\IEEEauthorblockN{Zongyang Ma}
\IEEEauthorblockA{\textit{School of AI and Advanced Computing}\\
\textit{Xi'an Jiaotong-Liverpool University} \\
mzyone@gmail.com}
\linebreakand
\IEEEauthorblockN{Fanyuan Meng}
\IEEEauthorblockA{\textit{Alibaba Business School}\\
\textit{Hangzhou Normal University} \\
fanyuan.meng@hotmail.com}
\and
\IEEEauthorblockN{Yan Xiao}
\IEEEauthorblockA{\textit{College of Information Engineering}\\
\textit{Shanghai Maritime University} \\
xiaoyan@shmtu.edu.cn}
\and
\IEEEauthorblockN{Fahui Miao}
\IEEEauthorblockA{\textit{College of Information Engineering}\\
\textit{Shanghai Maritime University} \\
miaofahui@126.com}
\linebreakand
\IEEEauthorblockN{Pearl Pu}
\IEEEauthorblockA{\textit{School of Computer and Communication Sciences}\\
\textit{\'{E}cole Polytechnique F\'{e}d\'{e}rale de Lausanne} \\
pearl.pu@epfl.ch}
}

\maketitle

\thispagestyle{plain}
\pagestyle{plain}

\begin{abstract}
Commonsense knowledge is essential for advancing natural language processing (NLP) by enabling models to engage in human-like reasoning, which requires a deeper understanding of context and often involves making inferences based on implicit external knowledge. This paper explores the integration of commonsense knowledge into various NLP tasks. We begin by reviewing prominent commonsense knowledge bases and then discuss the benchmarks used to evaluate the commonsense reasoning capabilities of NLP models, particularly language models. Furthermore, we highlight key methodologies for incorporating commonsense knowledge and their applications across different NLP tasks. The paper also examines the challenges and emerging trends in enhancing NLP systems with commonsense reasoning. All literature referenced in this survey can be accessed via our GitHub repository: \href{https://github.com/yuboxie/awesome-commonsense}{https://github.com/yuboxie/awesome-commonsense}.
\end{abstract}


\section{Introduction}
\IEEEPARstart{N}{atural} language processing (NLP) is a field of artificial intelligence (AI) that focuses on the interaction between computers and humans through natural language. The field of NLP has a rich history, dating back to the 1950s with the development of early machine translation systems~\cite{weaver1952translation,DBLP:conf/amta/Hutchins04}. Over the decades, NLP has evolved significantly, incorporating advancements in computational linguistics, artificial intelligence, and deep learning. These advancements have enabled the development of sophisticated NLP applications, such as speech recognition, sentiment analysis, and dialogue generation, which are now integral to many technologies we use daily. NLP encompasses a wide array of tasks, ranging from lexical and syntactical activities such as tokenization and part-of-speech tagging, to semantic and pragmatic challenges like reference resolution and text generation. While lexical and syntactical tasks are relatively straightforward and have largely been automated, higher-level semantic and pragmatic tasks demand a deeper understanding of natural language and remain challenging for machines. These tasks often require machines to reason using commonsense knowledge, which is universally accepted by humans but usually implicitly stated. For instance, if a six-foot-tall person is holding a two-foot-tall person and they are identified as father and son, it is intuitive to infer that the taller individual is the father~\cite{DBLP:journals/cacm/DavisM15}. This inference, obvious to humans, is difficult for machines because it relies on external commonsense knowledge that ``normally, a father is taller than a baby son.''

Since the 1960s, numerous efforts have been made to imbue NLP systems with commonsense reasoning abilities. Bar-Hillel~\cite{DBLP:journals/ac/Bar-Hillel60} was among the first to emphasize the importance of incorporating commonsense knowledge into NLP systems in the context of machine translation. Subsequent research has introduced various NLP tasks aimed at evaluating intelligent systems' commonsense reasoning capabilities. Broadly, these tasks, known as benchmarks, address two main types of commonsense knowledge: \textbf{naive physics} and \textbf{intuitive psychology}. Naive physics involves understanding how physical objects interact; for example, inferring that a glass of water falling to the floor will likely result in the glass shattering and the floor becoming wet. Intuitive psychology, on the other hand, enables us to infer people's behaviors, intents, or emotions, such as understanding that a person who has lost their job likely feels upset.

Commonsense knowledge is crucial for many real-world applications of NLP. For instance, virtual assistants like Siri and Alexa need to understand and respond appropriately to a wide range of user queries, often relying on commonsense reasoning. In online customer service, chatbots must handle diverse interactions, where they need to comprehend and address user issues effectively, often by inferring implicit context~\cite{DBLP:journals/access/Ait-MloukJ20,DBLP:series/asc/AugelloPVG14}. In sentiment analysis, NLP systems use commonsense knowledge to better understand the underlying emotions and sentiments in text, which helps in accurately interpreting the tone and context of user-generated content, such as reviews and social media posts~\cite{DBLP:conf/emnlp/ZhongWM19,DBLP:conf/acii/SureshO21,DBLP:conf/emnlp/GhosalMGMP20,nie2023long}. These applications demonstrate the importance of integrating commonsense reasoning to enhance the functionality and accuracy of NLP systems.

The most straightforward approach to enabling machines to reason with commonsense knowledge is to manually create comprehensive knowledge bases that encompass human commonsense knowledge. However, this task is immensely challenging due to the vastness and complexity of human knowledge, and the difficulty in determining which knowledge is considered common. Despite these challenges, significant efforts have been made to develop commonsense knowledge resources focused on specific domains, with the ultimate goal of creating a complete repository of human commonsense knowledge. One notable initiative in this area is the Open Mind Common Sense (OMCS)~\cite{DBLP:conf/coopis/SinghLMLPZ02} project, which aims to crowdsource commonsense knowledge from volunteers around the world. OMCS collects a wide range of commonsense facts and assertions, which are then used to build machine-readable knowledge bases such as ConceptNet~\cite{DBLP:conf/aaai/SpeerCH17}. These resources provide a valuable foundation for enhancing the commonsense reasoning capabilities of NLP systems.

Acquiring and utilizing commonsense knowledge in NLP is fraught with challenges. One major difficulty lies in encoding the implicit knowledge that humans take for granted. Additionally, commonsense can vary widely across different cultures and contexts, complicating the creation of universally applicable systems. Moreover, creating dynamic systems that can adapt to new information and contexts remains a significant hurdle. Emerging trends in the field of commonsense reasoning in NLP include the use of neural networks, deep learning, and large language models (LLMs) such as the GPT series~\cite{radford2018improving,radford2019language,DBLP:conf/nips/BrownMRSKDNSSAA20,DBLP:journals/corr/abs-2303-08774}. These technologies have shown promise in capturing and utilizing commonsense knowledge more effectively. Future advancements may include more sophisticated models capable of dynamic reasoning, integration of multimodal information, and the development of more comprehensive and nuanced commonsense knowledge bases.

This survey paper aims to provide an overview of the current state of commonsense reasoning in NLP. We review prominent commonsense knowledge bases and various benchmarks used to assess machines' commonsense reasoning abilities. Additionally, we highlight methodological research that leverages external commonsense knowledge resources to enhance general NLP tasks. Finally, we discuss potential future directions to further advance commonsense reasoning in NLP.

\section{Commonsense Knowledge Bases}
\begin{table*}[htbp]
    \small
    \renewcommand{\arraystretch}{1.3}
    \caption{Overview of prominent commonsense knowledge bases}
    \label{tab:knowledge_bases}
    \centering
    \begin{tabular}{|l||l|l|l|l|l|}
        \hline
        \thead{Name} & \thead{Type} & \thead{Size} & \thead{Form} & \thead{Language} & \thead{Creation} \\
        \hline
        OpenCyc~\cite{lenat1989building} & Naive physics & \makecell*[cl]{239,000 concepts\\2,039,000 facts} & Fact & English & Manual \\
        \hline
        ConceptNet~\cite{DBLP:conf/aaai/SpeerCH17} & \makecell*[cl]{Naive physics\\Intuitive psychology} & \makecell*[cl]{8 million nodes\\21 million links} & Graph & Multilingual & Crowdsourcing \\
        \hline
        SenticNet~\cite{DBLP:conf/cikm/CambriaLXPK20} & \makecell*[cl]{Intuitive psychology} & 200,000 concepts & Concept & Multilingual & Automatic \\
        \hline
        WebChild~\cite{DBLP:conf/acl/TandonMW17} & \makecell*[cl]{Naive physics\\Intuitive psychology} & \makecell*[cl]{2 million concepts\\18 million assertions} & Assertion & English & Automatic \\
        \hline
        EventKG~\cite{DBLP:conf/esws/GottschalkD18} & Naive physics & \makecell*[cl]{690,000 events\\2.3 million relations} & Graph  & Multilingual & Automatic \\
        \hline
        \textsc{Atomic}~\cite{DBLP:conf/aaai/SapBABLRRSC19} & \makecell*[cl]{Naive physics\\Intuitive psychology} & \makecell*[cl]{309,515 nodes\\877,108 triples} & Graph & English & Crowdsourcing \\
        \hline
        \textsc{Atomic}$_\text{20}^\text{20}$~\cite{DBLP:conf/aaai/HwangBBDSBC21} & \makecell*[cl]{Naive physics\\Intuitive psychology} & 1.33 million tuples & Graph & English & Crowdsourcing \\
        \hline
        ASER~\cite{DBLP:conf/www/ZhangLPSL20} & \makecell*[cl]{Naive physics\\Intuitive psychology} & \makecell*[cl]{194,000,677 nodes\\64,351,959 relations} & Graph & English & Automatic \\
        \hline
        GLUCOSE~\cite{DBLP:conf/emnlp/MostafazadehKMB20} & \makecell*[cl]{Naive physics\\Intuitive psychology} & 670,000 annotations & Annotation & English & Crowdsourcing \\
        \hline
        \textsc{Social-Chem-101}~\cite{DBLP:conf/emnlp/ForbesHSSC20} & Intuitive psychology & 292,000 rules & Rule & English & Crowdsourcing \\
        \hline
        CSKG~\cite{DBLP:conf/esws/IlievskiSZ21} & \makecell*[cl]{Naive physics\\Intuitive psychology} & \makecell*[cl]{2,160,968 nodes\\6,001,531 edges} & Graph & English & Automatic \\
        \hline
        MickeyCorpus~\cite{DBLP:conf/acl/LinLQ020} & \makecell*[cl]{Naive physics\\Intuitive psychology} & 561,000 sentences & Assertion & Multilingual & Automatic \\
        \hline
        \textsc{SocialDial}~\cite{DBLP:conf/sigir/ZhanLWLFKHQSSZS23} & Intuitive psychology & 6,433 dialogues & Dialogue & Chinese & Crowdsourcing \\
        \hline
        \textsc{NormDial}~\cite{DBLP:conf/emnlp/LiSSCM23} & Intuitive psychology & 4,231 dialogues & Dialogue & Chinese \& English & Crowdsourcing \\
        \hline
        \textsc{Moral Events}~\cite{zhang-etal-2024-moka} & Intuitive psychology & 5,494 annotations & Annotation & English & Crowdsourcing \\
        \hline
        COKE~\cite{coke2024} & Intuitive psychology & \makecell*[cl]{62,328 nodes\\45,369 cognitive chains} & Graph & English & \makecell*[cl]{Automatic \&\\Crowdsourcing} \\
        \hline
    \end{tabular}
\end{table*}

In this section, we give an overview of well-known commonsense knowledge bases. Different from lexical knowledge bases (such as WordNet~\cite{DBLP:journals/cacm/Miller95}) and factual knowledge bases (such as Wikidata~\cite{DBLP:journals/cacm/VrandecicK14}) widely used in natural language processing tasks, commonsense knowledge bases encode knowledge that is usually implicitly stated and considered obvious to most humans. Many commonsense knowledge bases are structured as \emph{knowledge graphs} in the form of nodes and edges, where nodes represent entities, and edges represent relationships between entities. Others are in the form of a set of \emph{facts/sentences}, expressed in propositional logic (used for symbolic inference) or natural language (used for language model training). Table~\ref{tab:knowledge_bases} gives an overview of well-known commonsense knowledge bases. Here, we are going to briefly introduce these knowledge bases one by one in their chronological order, denoting the year when its development began and the year when its newest version was published:

\begin{itemize}
    \item \textbf{Cyc (1984--2012)}. Cyc~\cite{lenat1989building} is an artificial intelligence project aiming at integrating ontologies and commonsense knowledge from all different domains into one knowledge base, and based on that, achieving the ability of knowledge inference like human beings. Concepts in Cyc are called ``constants'' and categorized into \emph{individuals}, \emph{collections}, \emph{truth functions}, and \emph{functions}. The Cyc project also includes an inference engine, which is capable of performing general logical deduction. Currently there are two releases of Cyc. OpenCyc 4.0 is the most recent public version and contains 239,000 concepts and 2,039,000 facts. ResearchCyc is licensed for research purposes and contains 500,000 concepts and 5,000,000 facts.

    \item \textbf{ConceptNet (1999--2020)}. ConceptNet~\cite{DBLP:conf/aaai/SpeerCH17} is a semantic network created by the Open Mind Common Sense (OMCS)~\cite{DBLP:conf/coopis/SinghLMLPZ02}, an artificial intelligence project aiming at building a large-scale commonsense knowledge base from the contributions of online users. It is a directed graph whose nodes are concepts, and the edges represent assertions of commonsense about the concepts, e.g., ``is a'', ``is used for'', ``motivated by goal'', etc. The nodes are natural language phrases, e.g., noun phrases, verb phrases, or clauses. ConceptNet contains over 8 million nodes and over 21 million links. The latest version of the knowledge base is ConceptNet 5.8.

    \item \textbf{SenticNet (2009--2024)}. As a knowledge base, SenticNet~\cite{DBLP:conf/cikm/CambriaLXPK20,DBLP:conf/lrec/CambriaLDXK22,cambria2024senticnet} provides a set of semantics, sentics, and polarity associated with 200,000 natural language concepts. Specifically, semantics define the denotative information associated with natural language phrases, sentics define the emotion categorization values (expressed in terms of four affective dimensions) associated with these concepts, and polarity is floating number between $-1$ and $+1$. The knowledge base is automatically created from multiple other resources, e.g., WordNet-Affect~\cite{DBLP:conf/lrec/StrapparavaV04} and OMCS. SenticNet also functions as an advanced AI framework comprising a suite of tools and methodologies for sentiment analysis, integrating commonsense reasoning, semiotics, psychology, linguistics, and machine learning. The latest iteration, SenticNet 8~\cite{cambria2024senticnet}, is a neurosymbolic AI framework that integrates commonsense knowledge representation with hierarchical attention networks. It surpasses traditional methods like bag-of-words, word2vec~\cite{DBLP:conf/nips/MikolovSCCD13}, RoBERTa~\cite{DBLP:journals/corr/abs-1907-11692}, and ChatGPT~\cite{RAY2023121} in accuracy, and is distinguished by its full interpretability, trustworthiness, and explainability.

    \item \textbf{WebChild (2014--2017)}. WebChild~\cite{DBLP:conf/acl/TandonMW17} is a large-scale commonsense knowledge base that was automatically extracted and disambiguated from Web contents, using semi-supervised label propagation over graphs of noisy candidate assertions. The knowledge base contains triples that connect nouns with adjectives via fine-grained relations like ``hasShape'', ``hasTaste'', ``evokesEmotion'', etc. The arguments of these assertions, nouns and adjectives, are disambiguated by mapping them onto their proper WordNet senses. The newest version WebChild 2.0 was released in 2017 and contains over 2 million concepts with 18 million assertions about them.
    
    \item \textbf{EventKG (2018)}. EventKG~\cite{DBLP:conf/esws/GottschalkD18} is a multilingual, event-centric temporal knowledge graph, that accentuates the events and temporal relationships. It encapsulates over 690,000 historical and contemporary events and in excess of 2.3 million temporal relations, structured through a canonical representation. The graph innovatively introduces \texttt{eventKG-s:Relation}, a class that interlinks two \texttt{sem:Core} instances---representing events or entities---to model temporal relationships between them, exemplified by relations such as ``Barack Obama"-``participated in"-``Second Inauguration of Barack Obama", enriched with annotations like validity time and \texttt{sem:RoleType} for comprehensive relation characterization.

    \item \textbf{\textsc{Atomic} (2019)}. \textsc{Atomic}~\cite{DBLP:conf/aaai/SapBABLRRSC19} is a comprehensive commonsense knowledge graph comprising 877K textual descriptions of inferential knowledge derived through crowdsourcing. This knowledge graph emphasizes \emph{if-then} relationships between events and the possible inferences that can be drawn from them. Specifically, it includes three types of relations: ``If-Event-Then-Mental-State'', ``If-Event-Then-Event'', and ``If-Event-Then-Persona''. The base events are extracted from diverse corpora, including stories and books. Overall, the \textsc{Atomic} knowledge graph encompasses 309,515 nodes and 877,108 If-Event-Then-* triples, providing a robust resource for understanding and modeling inferential reasoning.

    \item \textbf{\textsc{Atomic}$_\text{20}^\text{20}$ (2020)}. \textsc{Atomic}$_\text{20}^\text{20}$~\cite{DBLP:conf/aaai/HwangBBDSBC21} enhances the original \textsc{Atomic} by expanding the knowledge graph's scope and scale, incorporating a broader range of events and more complex relational structures. As a comprehensive commonsense knowledge graph, \textsc{Atomic}$_\text{20}^\text{20}$ comprises 1.33 million tuples spanning 23 commonsense relations, encompassing social, physical, and eventive aspects of everyday inferential knowledge. Studies have shown that pre-trained language models (PLMs) retrained on \textsc{Atomic}$_\text{20}^\text{20}$ can articulate knowledge more precisely than models trained solely on language data. Additionally, \textsc{Atomic}$_\text{20}^\text{20}$ serves as an effective training set for adapting language models, improving their ability to capture and express commonsense knowledge.

    \item \textbf{ASER (2020)}. ASER (activities, states, events, and their relations)~\cite{DBLP:conf/www/ZhangLPSL20} is a large-scale eventuality knowledge graph automatically extracted from more than 11-billion-token unstructured textual data. It contains 15 relation types belonging to five categories, 194 million unique eventualities, and 64 million edges between them. The eventualities were extracted from a wide range of corpora from different sources, according to a selected set of eventuality patterns. The eventuality relations were also automatically extracted using a selected set of seed connectives.

    \item \textbf{GLUCOSE (2020)}. The GLUCOSE~\cite{DBLP:conf/emnlp/MostafazadehKMB20} dataset comprises approximately 670,000 annotations that encode causal explanations grounded in narrative contexts, focusing on events, motivations, and emotional states. GLUCOSE is structured into ten dimensions of causal explanation, each providing a semi-structured format to capture both story-specific and generalized inference rules. The dataset is collected through a robust crowdsourcing platform, ensuring high-quality and diverse contributions from non-expert annotators. Unlike existing resources such as ConceptNet and \textsc{Atomic}, GLUCOSE offers broader contextual coverage, enabling models to generate nuanced and contextually appropriate commonsense inferences. Empirical evaluations demonstrate that models fine-tuned on GLUCOSE data significantly outperform pre-trained models in generating commonsense explanations. This dataset is poised to enhance AI applications in natural language understanding by facilitating models that more accurately mimic human-like reasoning across a range of narrative-driven tasks.

    \item \textbf{\textsc{Social-Chem-101} (2020)}. \textsc{Social-Chem-101}~\cite{DBLP:conf/emnlp/ForbesHSSC20} is a large-scale corpus designed to enhance language models' comprehension of the intents and underlying causes in human narratives. The corpus catalogs 292K descriptive rules-of-thumb (RoTs) and each RoT is further broken down with 12 different dimensions of people's judgments, including social judgments of good and bad, moral foundations, expected cultural pressure, and assumed legality, which together amount to over 4.5 million annotations of categorical labels and free-text descriptions. Each RoT within this collection is a descriptive cultural norm structured as the judgment of an action.

    \item \textbf{CSKG (2021)}. The CommonSense Knowledge Graph (CSKG)~\cite{DBLP:conf/esws/IlievskiSZ21} integrates seven distinct sources of commonsense knowledge into a unified graph to enhance AI systems' reasoning capabilities. CSKG addresses challenges in combining diverse knowledge sources, such as different modeling approaches and sparse overlap, by following five principles: embracing node heterogeneity, reusing edge types, leveraging external links, generating high-quality probabilistic links, and ensuring easy access to labels. It incorporates resources like ConceptNet, \textsc{Atomic}, Visual Genome~\cite{DBLP:journals/ijcv/KrishnaZGJHKCKL17}, and WordNet, creating a hyper-relational graph structure that improves connectivity and reasoning capabilities. CSKG offers comprehensive coverage with over 2.2 million nodes and 6 million edges, significantly increasing the availability of evidence for commonsense reasoning tasks. This integration allows for improved commonsense question answering and language model pre-training, offering substantial advancements over existing resources by providing enriched, interconnected commonsense knowledge.

    \item \textbf{MickeyCorpus (2021)}. The MickeyCorpus~\cite{DBLP:conf/acl/LinLQ020} is a multilingual dataset created to evaluate and enhance the performance of multilingual language models (ML-LMs) in commonsense reasoning tasks. Comprising 561,000 sentences across 11 languages, the MickeyCorpus serves as a resource for probing ML-LMs using the \textsc{MickeyProbe} task, which assesses the models' ability to rank sentences based on commonsense plausibility in a zero-shot setting. This corpus addresses the limitations of previous methods by providing a language-agnostic framework that accommodates multi-token concepts and ensures fair comparison across languages. Additionally, MickeyCorpus is integral to the proposed multilingual contrastive pre-training (MCP) strategy, which substantially improves the sentence-level representation of ML-LMs, leading to enhanced cross-lingual commonsense reasoning performance. By offering a comprehensive and diverse linguistic dataset, MickeyCorpus facilitates the development of more robust and culturally adaptable ML-LMs, advancing the field of natural language understanding beyond English.
    
    \item \textbf{\textsc{SocialDial} (2023)}. \textsc{SocialDial}~\cite{DBLP:conf/sigir/ZhanLWLFKHQSSZS23} is a dialogue corpus designed to enhance the development of socially-aware dialogue systems by incorporating Chinese social norms. It addresses the gap in existing dialogue systems that lack the ability to understand and integrate social norms. The dataset consists of 1,563 human-written multi-turn dialogues and 4,870 synthetic dialogues generated by ChatGPT, annotated with fine-grained social factors such as social relations, context, distance, and norms. \textsc{SocialDial} also serves as a benchmark for socially-aware dialogue systems. It includes an ontology-based synthetic data generation framework, which enables scalable and cost-effective creation of synthetic dialogues. Evaluations using state-of-the-art models like BERT~\cite{DBLP:conf/naacl/DevlinCLT19} and RoBERTa demonstrate the dataset's potential to enhance the performance of dialogue systems by incorporating social norms, thus providing a promising avenue for advancing socially-aware NLP applications.

    \item \textbf{\textsc{NormDial} (2023)}. \textsc{NormDial}~\cite{DBLP:conf/emnlp/LiSSCM23} is a synthetically generated bilingual (Chinese and English) dyadic dialogue dataset designed to study social norm adherence and violation within conversational contexts across Chinese and American cultures. Developed using a human-in-the-loop framework, the dataset consists of 4,231 dialogues with 29,550 conversational turns, each annotated for adherence or violation of social norms. These norms were initially defined by expert annotation and then expanded using LLMs to reflect both cultural contexts. The dialogues were evaluated and compared against existing datasets and were found to be of high quality, with strong naturalness and coherence ratings, although the models struggled with accurately identifying norm violations. The resource highlights the complexities of cross-cultural communication and provides a foundation for further research into developing systems capable of understanding nuanced social interactions across cultures.

    \item \textbf{\textsc{Moral Events} (2024)}. \textsc{Moral Events}~\cite{zhang-etal-2024-moka} is a dataset that encompasses annotations of news articles to analyze how different media ideologies report moral events, which are defined as events with moral implications, rooted in Moral Foundations Theory (MFT)~\cite{haidt2007morality,graham2009liberals,graham2013moral}, which includes polarities such as Care/Harm and Fairness/Cheating. These events are characterized by their moral evaluations, which arise when an agent with moral agency affects a patient who can be helped or harmed by the action. The dataset was curated using the moral event extraction (MEE) framework, which identifies morality-bearing event triggers, extracts participant entities, and infers the underlying moralities within unstructured texts.

    \item \textbf{COKE (2024)}. COKE~\cite{coke2024} is a COgnitive KnowledgE graph that aims to empower AI systems with Theory of Mind (ToM)~\cite{premack1978does} capabilities by formalizing ToM as a comprehensive collection of over 45,000 manually verified cognitive chains. These chains encapsulate human mental activities and their corresponding behavioral and emotional responses within specific social contexts. Composed of five node types---situations, clues, thoughts, actions, and emotions---COKE forms cognitive chains labeled with either positive or negative polarity. To overcome the limitation of addressing all real-world scenarios, COKE integrates with the cognitive language model COLM, which combines commonsense knowledge from LLMs with ToM capabilities to infer cognitive chains for novel situations. Experimental results demonstrate that COLM significantly outperforms baseline models like GPT-4~\cite{DBLP:journals/corr/abs-2303-08774} in generating cognitive chains and enhancing applications such as emotional support conversations, highlighting the potential of integrating cognitive reasoning into AI systems.
\end{itemize}

\section{Benchmarks}
\begin{figure*}
    \centering
    \includegraphics[width=\textwidth]{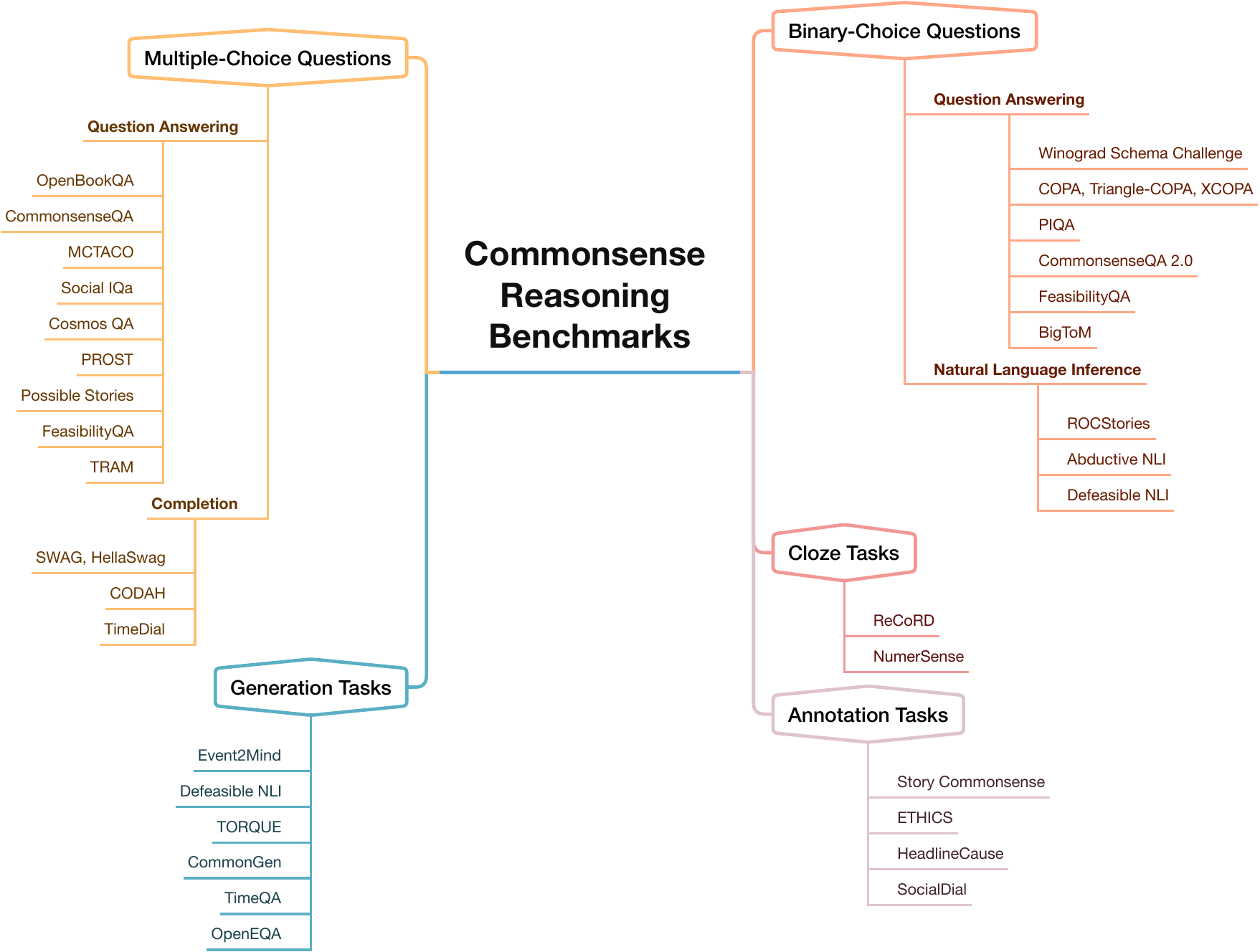}
    \caption{Categorization of commonsense reasoning benchmarks by task type.}
    \label{fig:benchmark_categories}
\end{figure*}

\begin{figure*}
    \centering
    \includegraphics[width=0.75\textwidth]{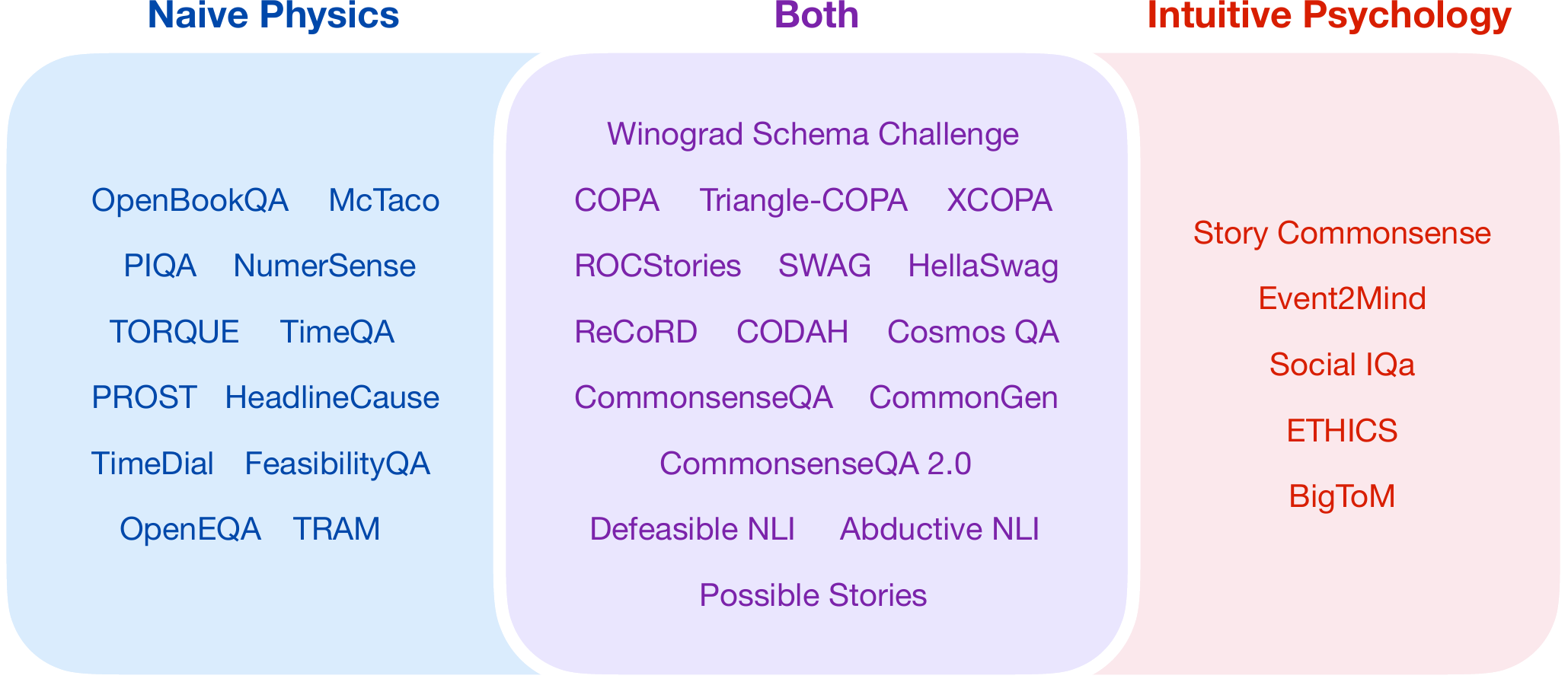}
    \caption{Overview of commonsense reasoning benchmarks categorized by the types of commonsense knowledge they evaluate.}
    \label{fig:benchmark_knowledge_type}
\end{figure*}

Advanced NLP tasks, such as machine reading comprehension and natural language inference, present significant challenges as they require systems to emulate human-like reasoning and draw inferences based on commonsense knowledge. While the objectives of these tasks differ, successfully addressing them typically necessitates a degree of commonsense reasoning within the systems. To evaluate the commonsense reasoning capabilities of NLP models, particularly LLMs, numerous benchmarks have been developed. In this section, we review prominent benchmarks that demand commonsense reasoning from models. We categorize these benchmarks into several types, including \emph{multiple-choice questions}, \emph{binary-choice questions}, \emph{classification tasks}, \emph{cloze tasks}, and \emph{generation tasks}, as illustrated in Fig.~\ref{fig:benchmark_categories}. Additionally, we classify these benchmarks based on the specific types of commonsense knowledge they assess, as depicted in Fig.~\ref{fig:benchmark_knowledge_type}. The following subsections provide a brief introduction to these benchmarks according to their task type.

\subsection{Multiple-Choice Questions}
The majority of benchmarks are structured as multiple-choice questions, where the evaluated model is presented with several options and must select the correct one or more. In one common scenario, referred to as \emph{question answering}, the benchmark poses a specific question, and the model is required to identify the correct answer(s) from the given choices. Some benchmarks also include supplementary text, often in the form of a story or relevant facts, to provide context alongside the question. Prominent benchmarks within this category include:
\begin{itemize}
    \item \textbf{OpenBookQA}~\cite{DBLP:conf/emnlp/MihaylovCKS18} is a benchmark dataset designed to assess the ability of AI systems to combine core scientific knowledge with broader commonsense reasoning in answering elementary-level science questions. The dataset, consisting of nearly 6,000 multiple-choice questions, requires models to integrate provided science facts with additional external knowledge to solve problems that mimic the open-book exam setting. Unlike traditional question-answering datasets, OpenBookQA emphasizes the need for multi-hop reasoning and effective knowledge retrieval, making it a more challenging task.

    \item \textbf{\textsc{CommonsenseQA}}~\cite{DBLP:conf/naacl/TalmorHLB19} is a dataset designed to evaluate the capacity of natural language understanding systems to perform commonsense reasoning. Derived from the ConceptNet knowledge graph, the dataset consists of 12,247 multiple-choice questions and requires reasoning that extends beyond simple fact retrieval, often necessitating the application of background knowledge. Each question in CommonsenseQA is carefully constructed to present contextually complex scenarios, making it more difficult for models to perform accurately.

    \item \textbf{\textsc{McTaco}} (short for multiple choice temporal commonsense)~\cite{DBLP:conf/emnlp/ZhouKNR19} is a dataset designed to evaluate temporal commonsense reasoning in natural language understanding. It systematically addresses five key temporal properties: duration, temporal ordering, typical time, frequency, and stationarity. Created via crowdsourcing, \textsc{McTaco} provides a challenging test set for evaluating the performance of NLP models, such as BERT~\cite{DBLP:conf/naacl/DevlinCLT19} and ESIM~\cite{DBLP:conf/acl/ChenZLWJI17}, on tasks requiring temporal reasoning.

    \item \textbf{\textsc{Social IQa}}~\cite{DBLP:conf/emnlp/SapRCBC19} is a large-scale benchmark for evaluating and improving NLP models' social and emotional intelligence through commonsense reasoning about social interactions. It includes 38,000 multiple-choice questions designed to challenge models in understanding motivations, emotional reactions, and outcomes of everyday scenarios. By using a crowdsourcing method that reduces biases in incorrect answers, \textsc{Social IQa} offers a robust dataset for training NLP models. Despite advancements, models like BERT still fall short of human performance on these tasks, but \textsc{Social IQa} has proven effective for enhancing performance on related commonsense reasoning challenges such as the Winograd Schema Challenge~\cite{DBLP:conf/aaaiss/Levesque11} and COPA~\cite{DBLP:conf/aaaiss/RoemmeleBG11}.

    \item \textbf{\textsc{Cosmos QA}}~\cite{DBLP:conf/emnlp/HuangBBC19} is a comprehensive machine reading comprehension dataset designed to challenge models with the need for contextual commonsense reasoning. The dataset comprises 35,588 multiple-choice questions based on a diverse range of everyday situations drawn from personal narratives. Unlike traditional reading comprehension datasets, \textsc{Cosmos QA} emphasizes reasoning that goes beyond the explicit text, requiring models to infer causes, effects, and hypothetical scenarios.

    \item \textbf{PROST} (Physical Reasoning about Objects through Space and Time)~\cite{DBLP:conf/acl/Aroca-Ouellette21} is a probing dataset designed to evaluate the capability of language models to reason about physical interactions in the real world. Comprising 18,736 multiple-choice questions generated from 14 manually curated templates across 10 distinct physical reasoning concepts, PROST challenges models in a zero-shot setting, emphasizing the limitations of current state-of-the-art pre-trained models. The results demonstrate that these models often rely on spurious correlations, such as the order of answer options, rather than a genuine understanding of physical concepts like direction, mass, and breakability. PROST highlights the significant gap between human-like physical reasoning and the performance of existing language models, underscoring the necessity for multimodal training that incorporates real-world experiences to enhance model capabilities in understanding and reasoning about physical phenomena.

    \item \textbf{Possible Stories}~\cite{DBLP:conf/coling/AshidaS22} provides a comprehensive evaluation framework for situated commonsense reasoning, focusing on scenarios where multiple plausible outcomes are possible. The benchmark consists of 4,533 multiple-choice questions based on 1,313 story passages, each designed to assess the ability of language models to reason about potential outcomes in varied contexts. The dataset includes diverse scenarios that require counterfactual reasoning and an understanding of characters' motivations, emotions, and fictional contexts.

    \item \textbf{FeasibilityQA}~\cite{DBLP:conf/eacl/GuptaVMPSSGB23} is a dataset designed to evaluate the ability of NLP models to reason about the feasibility of actions or their effects. This dataset includes binary classification (BCQ) and multi-choice multi-correct questions (MCQ) that test models on commonsense reasoning about feasibility, a fundamental yet underexplored aspect of natural language understanding. Despite the sophistication of language models like GPT-3~\cite{DBLP:conf/nips/BrownMRSKDNSSAA20}, GPT-2~\cite{radford2019language}, and T5~\cite{DBLP:journals/jmlr/RaffelSRLNMZLL20}, they demonstrate significant struggles with these tasks, achieving low accuracy rates (e.g., 19\% and 62\% in MCQ and BCQ, respectively, in zero-shot settings).

    \item \textbf{TRAM} (Temporal Reasoning for large lAnguage Models)~\cite{wang-zhao-2024-tram} serves as a comprehensive benchmark designed to evaluate the temporal reasoning capabilities of LLMs. It comprises ten diverse tasks, ranging from foundational temporal understanding (e.g., duration, frequency) to advanced temporal interpretations and computations (e.g., ambiguity resolution, arithmetic, causality). TRAM includes 526.7K questions across 38 subtasks, all formatted as multiple-choice questions to ensure consistency and ease of evaluation. The benchmark reveals that even the most advanced models, such as GPT-4, significantly lag behind human performance in temporal reasoning tasks.
\end{itemize}

In addition to the question-answering scenario, some benchmarks employ a \emph{completion} task, where the model is required either to select the option that best continues the given context or to choose the option that best fits into a blank space. Prominent benchmarks in this category include:

\begin{itemize}
    \item \textbf{SWAG} (Situations With Adversarial Generations)~\cite{DBLP:conf/emnlp/ZellersBSC18} provides 113K multiple-choice questions designed to predict the most plausible continuation of a given scenario. To address annotation artifacts and biases common in existing datasets, SWAG employs a novel Adversarial Filtering (AF) technique, which iteratively uses an ensemble of classifiers to remove stylistic biases, thereby producing a more challenging and robust dataset. \textbf{HellaSwag} (Harder Endings, Longer contexts, and Low-shot Activities for Situations With Adversarial Generations)~\cite{DBLP:conf/acl/ZellersHBFC19} expands on the SWAG benchmark by increasing the complexity and diversity of context through sources like WikiHow and ActivityNet, pushing the boundaries of what current models can achieve. It includes 70K dataset examples in total, with 25K best ActivityNet contexts (i.e., those with the highest agreement among crowd workers) and 45K best WikiHow contexts.

    \item \textbf{CODAH} (COmmonsense Dataset Adversarially-authored by Humans)~\cite{DBLP:journals/corr/abs-1904-04365} extends the SWAG dataset by introducing adversarially-constructed multiple-choice questions that target the weaknesses of state-of-the-art models such as BERT and GPT. Human annotators, incentivized to generate questions that confound these models, contributed to the creation of 2,801 questions. CODAH includes various question types, such as idioms, negation, and quantitative reasoning, and has proven to be a more difficult benchmark than SWAG.

    \item \textbf{\textsc{TimeDial}}~\cite{DBLP:conf/acl/QinGUHCF20} addresses the challenge of temporal commonsense reasoning in dialogues, focusing on the ability of language models to understand and reason about temporal concepts within multi-turn conversations. \textsc{TimeDial} introduces a novel multiple-choice blank-filling task with over 1,100 carefully curated dialogues requiring models to infer correct temporal expressions based on context, world knowledge, and arithmetic reasoning. Empirical evaluations reveal that LLMs like T5 and GPT-3 struggle with this task, significantly underperforming compared to human baselines. The analysis further shows that these models often rely on shallow text patterns rather than true contextual reasoning, highlighting the need for more robust approaches to model temporal commonsense in dialogues.
\end{itemize}

\subsection{Binary-Choice Questions}
Some benchmarks are designed as binary-choice questions, where the model must select the correct answer from only two options rather than multiple options, as is the case with multiple-choice questions. Consequently, these tasks are inherently easier due to the reduced number of choices. Similarly, a common example of this type of task is \emph{question answering}, with several notable benchmarks falling into this category, including:
\begin{itemize}
    \item The \textbf{Winograd Schema Challenge} (WSC)~\cite{DBLP:conf/aaaiss/Levesque11} is proposed as an alternative to the Turing Test~\cite{turing2009computing}, focusing on a machine's ability to resolve referential ambiguities that are trivial for humans but challenging for AI. A Winograd schema consists of a pair of sentences that differ by only one or two words, leading to a referential ambiguity that requires common sense reasoning and world knowledge to resolve. The WSC avoids pitfalls associated with statistical methods and simple linguistic tricks, making it a more robust measure of machine understanding. The challenge emphasizes the need for NLP systems to engage in deep reasoning, distinguishing it from other NLP tasks that rely heavily on statistical approaches. As such, the WSC represents a significant advancement in evaluating AI's capacity for genuine comprehension and intelligent behavior.

    \item \textbf{COPA} (Choice Of Plausible Alternatives)~\cite{DBLP:conf/aaaiss/RoemmeleBG11} is a benchmark that involves causal inference between events. The dataset comprises 1,000 examples, each presenting an event followed by a question asking the model to select the correct cause or effect from two options. \textbf{Triangle-COPA}~\cite{DBLP:conf/aaai/Gordon16} is a variation of COPA, containing 100 examples in the same format, but supplemented with videos depicting interactions between a circle and a triangle. The questions in Triangle-COPA focus more on emotions and intentions. However, scalability remains a challenge for both datasets when evaluating modern language models. To advance NLP tools in languages other than English and address the Anglocentric bias in commonsense reasoning models, \textbf{XCOPA}~\cite{DBLP:conf/emnlp/PontiGMLVK20} was introduced. This multilingual dataset supports causal commonsense reasoning across 11 languages and is typologically diverse.

    \item \textbf{PIQA} (Physical Interaction: Question Answering)~\cite{DBLP:conf/aaai/BiskZLGC20} addresses the challenge of physical commonsense reasoning in natural language understanding. While humans find physical reasoning intuitive, NLP models, such as BERT and RoBERTa~\cite{DBLP:journals/corr/abs-1907-11692}, struggle significantly with these tasks. PIQA introduces a benchmark dataset focused on physical interaction scenarios, requiring models to choose the most sensible solution from a pair of options. This benchmark aims to drive progress in developing models that better capture physical commonsense knowledge.

    \item \textbf{\textsc{CommonsenseQA} 2.0}~\cite{DBLP:conf/nips/TalmorYBBGCB21} introduces a gamified framework for data creation. By engaging users in a game where they generate challenging yes/no questions designed to mislead AI, this approach ensures the collection of diverse and difficult questions. The dataset, consisting of 14,343 examples, is challenging for existing models, with the best-performing model, \textsc{Unicorn}-11B~\cite{DBLP:conf/aaai/LourieBBC21}, achieving only 70.2\% accuracy, far below the 94.1\% human accuracy.

    \item \textbf{BigToM}~\cite{DBLP:conf/nips/GandhiFGG23} is a framework designed for evaluating the social reasoning capabilities of LLMs by procedurally generating Theory of Mind (ToM) evaluations. It addresses limitations in previous methodologies, such as ambiguous or insufficiently controlled test items, by using causal templates to create a new social reasoning benchmark consisting of 25 controls and 5,000 model-generated evaluations. The benchmark, evaluated and rated by human participants, was found to be superior to crowd-sourced tests and comparable to expert-written evaluations. BigToM's results indicate that while GPT-4 demonstrates ToM capabilities that mirror human inference patterns, its performance remains less reliable compared to humans, and other LLMs struggle significantly more. This framework offers a scalable, cost-efficient method for assessing and understanding the strengths and weaknesses of LLMs in ToM reasoning, providing valuable insights into the models' ability to simulate human-like social cognition.
\end{itemize}

In another scenario, the evaluated model performs a \emph{natural language inference (NLI)} task, where it must select the option that is entailed by the given text. Some benchmarks in this category include:
\begin{itemize}
    \item \textbf{ROCStories}~\cite{DBLP:conf/naacl/MostafazadehCHP16}, a corpus of 50,000 five-sentence commonsense stories, was developed to address the need for a robust evaluation framework in the understanding of causal and temporal relations in narratives. It is coupled with an evaluation framework that challenges models to select the correct ending for a four-sentence story from two alternatives. Experimental results reveal that models relying on shallow language understanding struggle with this task, emphasizing the need for richer semantic representations in story understanding and script learning.
    
    \item \textbf{Abductive NLI}~\cite{DBLP:conf/iclr/BhagavatulaBMSH20} focuses on the task of identifying the most plausible explanation for a given set of observations. It is supported by the ART dataset, which includes over 20,000 commonsense narrative contexts and 200,000 explanatory hypotheses. Despite achieving a 68.9\% accuracy with the best model, there is a substantial gap compared to human performance at 91.4\%.
    
    \item \textbf{Defeasible NLI}~\cite{DBLP:conf/emnlp/RudingerSHBFBSC20} introduces a dataset and corresponding tasks designed to explore the concept of defeasible inference within NLP. It extends existing inference datasets to incorporate scenarios where an initial inference might be either strengthened or weakened by additional contextual information. Two tasks were formulated: a classification task that determines whether a new piece of information strengthens or weakens an inference, and a generative task that requires creating such contextual updates. While LLMs perform well on the classification task, the generative task remains challenging.
\end{itemize}

\subsection{Cloze Tasks}
In a cloze task, the given context contains one or more blanks, and the evaluated model must fill in these blanks with appropriate words or phrases. Unlike in multiple-choice completion tasks, the model is not provided with a set of options but must predict the missing words based solely on the context or its training vocabulary. Two well-known benchmarks that utilize the cloze task are:
\begin{itemize}
    \item \textbf{ReCoRD} (Reading Comprehension with Commonsense Reasoning Dataset)~\cite{DBLP:journals/corr/abs-1810-12885} is a large-scale machine reading comprehension (MRC) dataset designed to evaluate systems' abilities to perform commonsense reasoning. Consisting of over 120,000 examples derived from news articles, ReCoRD requires models to not only match patterns in the text but also to engage in deep commonsense reasoning to infer answers. Unlike other datasets, which can often be addressed with surface-level text matching, ReCoRD's cloze-style queries demand an understanding that spans multiple sentences and incorporates implicit commonsense knowledge.

    \item \textbf{\textsc{NumerSense}}~\cite{DBLP:conf/emnlp/LinLKR20} explores the limitations of LLMs like BERT and RoBERTa in understanding and recalling numerical commonsense knowledge. Despite their success in various NLP tasks, these models perform poorly when required to predict masked numerical values in sentences, such as determining the number of legs a bird typically has. \textsc{NumerSense} introduces a diagnostic dataset of 13.6K examples, revealing that LLMs struggle with consistency and accuracy in numerical commonsense tasks, even after fine-tuning with distant supervision.
\end{itemize}

\subsection{Annotation Tasks}
In an annotation task, the evaluated model is required to label a given text using a predefined set of labels, often based on commonsense knowledge. This process is essentially a text classification task. Notable benchmarks are:
\begin{itemize}
    \item \textbf{Story Commonsense}~\cite{DBLP:conf/acl/KnightCSRB18} is a large-scale dataset containing detailed annotations that link commonsense story events to the mental states of characters, even when these states are not explicitly mentioned. It leverages theories from psychology, including Maslow's hierarchy of needs~\cite{maslow1943theory}, Reiss's basic desires~\cite{reiss2004multifaceted}, and Plutchik's wheel of emotions~\cite{plutchik1980general}, to categorize the motivations and emotional reactions of characters. The dataset, which covers over 15,000 stories and includes 300,000 low-level annotations, serves as a foundation for new tasks in natural language understanding, such as mental state tracking and explanation generation.

    \item \textbf{ETHICS}~\cite{DBLP:conf/iclr/HendrycksBBC0SS21} is a benchmark for evaluating language models' understanding of basic human ethical concepts across diverse scenarios. The dataset covers five major areas of normative ethics: justice, deontology, virtue ethics, utilitarianism, and commonsense morality. It challenges models to make morally informed decisions by connecting factual knowledge with value judgments in complex, open-world contexts. Current models show some ability to predict ethical judgments but are far from mastering the nuanced understanding required for ethical AI.

    \item \textbf{HeadlineCause}~\cite{DBLP:conf/lrec/GusevT22} is a dataset designed to detect implicit causal relations between pairs of news headlines, addressing the challenges in existing datasets that focus predominantly on either commonsense causal reasoning or explicit causal relations. Comprising over 5,000 headline pairs in English and 9,000 in Russian, this dataset was annotated via crowdsourcing and includes a variety of relationships, from unrelated headlines to those involving causation and refutation. The dataset is particularly notable for its emphasis on implicit, inter-sentence causal relations, which require models to leverage both commonsense and world knowledge.
\end{itemize}

\subsection{Generation Tasks}
In a generation task, the evaluated model is expected to generate the answer independently, rather than selecting from a set of predefined options. One approach to this is to reformulate the problem as a classification task, determining which words or n-grams should constitute the answer. A more common approach, however, is to use an autoregressive decoder that generates the answer token by token. Notable benchmarks in this category include:
\begin{itemize}
    \item \textbf{Event2Mind}~\cite{DBLP:conf/acl/SmithCSRA18} is a commonsense inference benchmark focusing on predicting the intents and reactions of participants in everyday events. The model utilizes a corpus of 25,000 event phrases, crowdsourced to include a diverse range of scenarios. By encoding event phrases and generating textual descriptions of intents and reactions, Event2Mind aims to understand the mental states associated with events.

    \item \textbf{TimeQA}~\cite{DBLP:conf/nips/ChenWWW21} is a dataset specifically designed to evaluate the ability of NLP models to handle time-sensitive queries that require temporal reasoning. The dataset was constructed by mining time-evolving facts from Wikidata~\cite{DBLP:journals/cacm/VrandecicK14}, aligning these facts with corresponding Wikipedia passages, and generating question-answer pairs based on these annotated time-sensitive facts. TimeQA presents significant challenges in both temporal understanding, where models must grasp the time scope of facts scattered across long documents, and temporal reasoning, where models need to infer and reason over temporal relationships.


    \item \textbf{\textsc{Torque}}~\cite{DBLP:conf/emnlp/NingWHPGR20} is a reading comprehension dataset specifically designed to evaluate a model's ability to understand temporal relationships between events in text. Comprising 3.2K news snippets and over 21K human-generated questions, \textsc{Torque} addresses a critical gap in current MRC benchmarks, which typically do not include questions that assess temporal reasoning. The dataset challenges models with complex queries about the sequence and timing of events, such as identifying what happened before or after a specific event, or what events occurred simultaneously.

    \item \textbf{CommonGen}~\cite{DBLP:conf/akbc/LinSZZBCR20} introduces a constrained text generation task designed to assess and advance generative commonsense reasoning in machines. The task requires models to generate coherent sentences that describe everyday scenarios using a specified set of common concepts, emphasizing the need for relational reasoning with background commonsense knowledge and compositional generalization to handle novel concept combinations. The accompanying dataset includes 77K sentences across 35K unique concept sets, offering a challenging benchmark for state-of-the-art language models.

    \item \textbf{OpenEQA}~\cite{Majumdar_2024_CVPR} is an open-vocabulary Embodied Question Answering (EQA)~\cite{DBLP:conf/cvpr/DasDGLPB18} benchmark that supports both episodic memory and active exploration scenarios. It evaluates an AI agent's ability to understand and interact with real-world environments to answer complex, natural language questions. OpenEQA includes over 1,600 human-generated questions based on 180 real-world environments, testing various cognitive capabilities such as spatial reasoning, object recognition, and world knowledge. The benchmark incorporates a novel evaluation protocol using LLMs to score responses, achieving strong alignment with human judgments. Despite the sophistication of current foundation models, including GPT-4V~\cite{DBLP:journals/corr/abs-2309-17421}, performance on OpenEQA lags significantly behind human-level accuracy, highlighting substantial challenges in developing agents capable of robust, real-world EQA.
\end{itemize}

\subsection{Multi-Task Benchmarks}
Some benchmarks consist of multiple tasks, and the performance of evaluated systems is typically measured by averaging their scores across these tasks. \textbf{GLUE} (General Language Understanding Evaluation)~\cite{DBLP:conf/iclr/WangSMHLB19} is a widely used comprehensive benchmark for evaluating NLP systems, featuring tasks that require commonsense reasoning, such as natural language inference, textual entailment, and question answering. \textbf{SuperGLUE}~\cite{DBLP:conf/nips/WangPNSMHLB19} builds upon GLUE by incorporating more challenging natural language understanding tasks, designed to meet the demands of increasingly advanced NLP systems in recent years. \textbf{CLUE}~\cite{DBLP:conf/coling/XuHZLCLXSYYTDLS20} (Chinese Language Understanding Evaluation) is a comprehensive benchmark developed to advance Chinese natural language processing. It addresses the gap left by predominantly English-centric benchmarks like GLUE and SuperGLUE by introducing a set of nine diverse language understanding tasks. \textbf{LOT} (LOng Text)~\cite{DBLP:journals/tacl/GuanFCHMFH22} is a story-centric benchmark for evaluating the understanding and generation of Chinese long texts. It includes two understanding tasks and two generation tasks, all based on human-written Chinese stories to test key abilities such as commonsense reasoning, inter-sentence relations, and coherence. \textbf{BIG-bench} (Beyond the Imitation Game benchmark)~\cite{DBLP:journals/tmlr/SrivastavaRRSAF23} is a large-scale, diverse, and challenging benchmark designed to evaluate the capabilities of current and future language models across a wide range of tasks. It includes 204 tasks contributed by over 450 authors from 132 institutions, covering topics from linguistics to social biases and software development.

\section{Methodologies}
This section explores and compares various methods for enhancing commonsense knowledge in NLP models. By discussing these methods, we aim to provide a comprehensive perspective on how different commonsense knowledge enhancement strategies are applied in NLP models, particularly in language models. Given that commonsense knowledge is a specialized form of knowledge, general knowledge enhancement methods are also applicable. Therefore, we adopt the methodology categorization proposed by Wang et al.~\cite{DBLP:journals/corr/abs-2310-16218}, with a focus on how commonsense knowledge is specifically enhanced in recent papers.

\begin{figure*}
    \centering
    \subfloat[]{\includegraphics[width=0.8\textwidth]{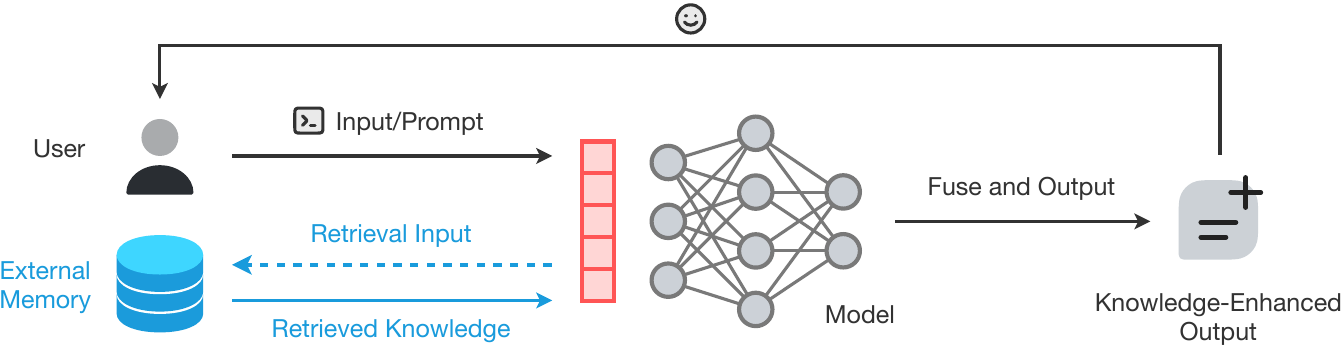}
    \label{fig:methodologies_external}}\\
    \subfloat[]{\includegraphics[width=0.77891037\textwidth]{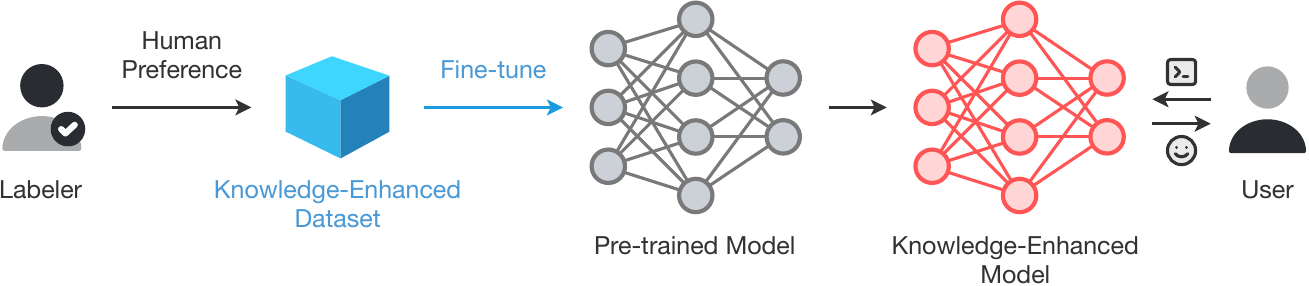}
    \label{fig:methodologies_global}}\\
    \subfloat[]{\includegraphics[width=0.77891037\textwidth]{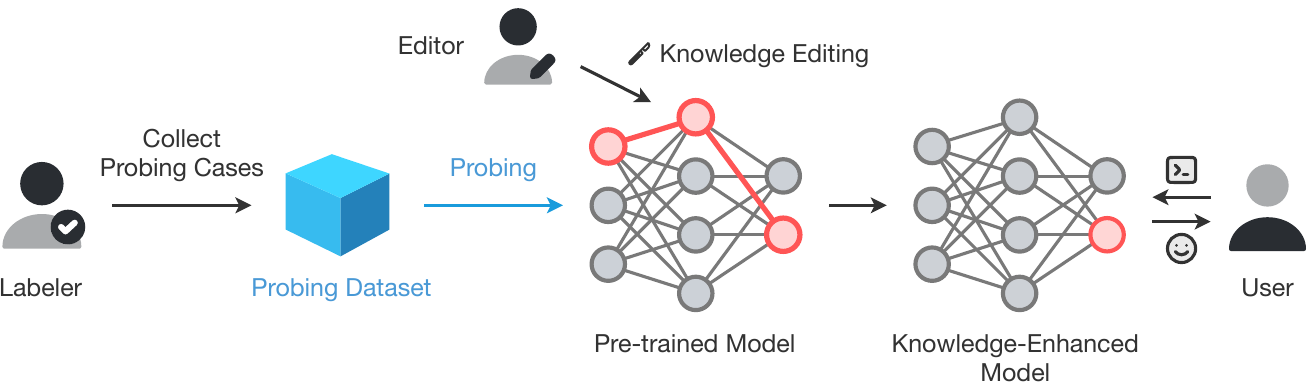}
    \label{fig:methodologies_local}}
    \caption{Three different ways for NLP models to incorporate and enhance commonsense knowledge: (a) External Memorization; (b) Global Optimization; (c) Local Modification.}
    \label{fig:methodologies}
\end{figure*}

\subsection{External Memorization}
\label{subsec:external_memorization}
Enhancing language models by incorporating commonsense knowledge through external retrieval is a popular and effective method for boosting their understanding, as shown in Fig.~\ref{fig:methodologies_external}. Under this category, commonsense knowledge enhancement techniques can be broadly classified into \emph{static knowledge retrieval} and \emph{dynamic knowledge generation}. Static knowledge retrieval methods depend on pre-existing knowledge bases and employ precise retrieval algorithms to extract pertinent information, whereas dynamic knowledge generation methods utilize specialized models to generate real-time commonsense reasoning that is relevant to the current context. In open-domain conversational systems, integrating commonsense knowledge is essential for generating responses that align more closely with human reasoning and provide deeper insights. Many recent studies have adopted the external memorization approach for this purpose. Next, we will review some of the key works in this area.

The commonsense knowledge aware conversational model (CCM) introduced by Zhou et al.~\cite{DBLP:conf/ijcai/ZhouYHZXZ18}, represents an initiative in integrating extensive commonsense knowledge into generating conversations in open domains. The model operates by fetching pertinent knowledge graphs from a knowledge base upon receiving a user prompt. Subsequently, it encodes these graphs utilizing a \textit{static graph attention mechanism} to enhance the semantic content of the input, thereby aiding in better comprehension. During the word generation phase, the model carefully analyzes the retrieved knowledge graphs and the knowledge triples contained within each graph to improve the generation process via a \textit{dynamic graph attention mechanism}. By employing this dual graph attention approach, CCM effectively captures and encodes more organized semantic details. Wu et al.~\cite{DBLP:conf/acl/WuLZZW20} found that previous commonsense knowledge-enhanced dialogue systems (e.g., CCM) had a key limitation: these systems usually retrieve knowledge facts without taking into account the specific dialogue context, which can result in the introduction of irrelevant knowledge facts (noises). To address this issue, they proposed commonsense knowledge-aware dialogue generation model (ConKADI), which consists of: a knowledge retriever that extracts relevant commonsense fact triples based on the query message, a context encoder that summarizes utterances into contextual representations, and a felicitous fact recognizer that calculates the probability distribution of relevant facts over the retrieved set. This distribution is then used to initialize the decoder and guide dialogue generation, employing a 0-1 indicator vector for supervised training. Ling et al.~\cite{DBLP:journals/corr/abs-2310-11672} extended the \textit{retrieve-then-select} approach to more general open-ended dialogue systems. Recognizing that open-domain commonsense reasoning often requires implicit multi-hop reasoning, they proposed the external KnowlEdge-Enhanced Prompting method (KEEP). This method constructs the local knowledge graph based on the context and conceptual knowledge graphs (e.g., ConceptNet~\cite{DBLP:conf/aaai/SpeerCH17}), and then incorporates the implicit knowledge in pre-trained language models to prune irreverent path. Cai et al.~\cite{DBLP:conf/acl/CaiSXSWGZX23} applied a similar methodology to enhance empathetic dialogue generation.

The following methods dynamically generate relevant commonsense knowledge, which is different from the methods mentioned above. Liu et al.~\cite{DBLP:journals/corr/abs-2302-01441} proposed a novel approach that uses the COMET-BART model~\cite{DBLP:conf/aaai/HwangBBDSBC21} to extract implicit commonsense knowledge from the dialogue context. To improve the language model's understanding of the obtained commonsense relations, they developed a method to express the commonsense tuples using natural language templates. These verbalized commonsense statements were then added to the dialogue history as \emph{additional input text}. Similarly, Finch et al.~\cite{DBLP:journals/corr/abs-2406-09138} proposed a method that employs the T5-based ConvoSenseGenerator~\cite{DBLP:journals/tacl/FinchC24} to produce inferences based on the dialogue context and specified commonsense types. Subsequently, GPT-3.5 is used to assess each inference, carefully selecting the most useful, relevant, and interesting ones for the next response. The final response generation step incorporates the selected inferences and the dialogue context as input to produce the next utterance. The dynamic heterogeneous-graph reasoning method with language models and knowledge representation learning (DHLK)~\cite{DBLP:conf/acl/WangZLL23} presents a novel approach to integrating dynamic commonsense knowledge into language models. It uses a two-stage pruning strategy that (i) filters key entities based on the dictionary vocabulary to achieve the first-stage pruning while incorporating the paraphrases in the dictionary into the subgraph to construct the heterogeneous knowledge graph (HKG), and (ii) encodes and fuses the question answering context and HKG using a language model, and dynamically removes irrelevant KG entities based on the attention weights of the language model for the second-stage pruning. Finally, DHLK performs answer reasoning on the HKG by relation mask self-attention (RMSA).

\subsection{Global Optimization}
As shown in Fig.~\ref{fig:methodologies_global}, global optimization methods integrate commonsense knowledge by first preparing a knowledge-enhanced dataset and then fine-tuning a pre-trained model (updating all of its parameters) to strengthen the model's grasp of the target knowledge. This fine-tuning process ensures the model achieves desirable results regarding the target knowledge, giving global optimization the advantages of precision and generalization. However, this approach can struggle to preserve the model's original knowledge, and fine-tuning is often time-consuming and prone to overfitting. Next, we review recent works on commonsense reasoning that employ the global optimization approach.

To enhance temporal commonsense reasoning, Kimura et al.~\cite{DBLP:conf/ijcnlp/KimuraPK22} proposed a language model created by multi-step fine-tuning, continual pre-training, and multi-task learning, using pre-trained models such as BERT~\cite{DBLP:conf/naacl/DevlinCLT19}, RoBERTa~\cite{DBLP:journals/corr/abs-1907-11692}, and ALBERT~\cite{DBLP:conf/iclr/LanCGGSS20}. Their approach leverages masked language modeling and the \textsc{McTaco} dataset~\cite{DBLP:conf/emnlp/ZhouKNR19} to predict temporal indicators, improving the performance on time-related tasks. Notably, their experiments demonstrate that ALBERT, when fine-tuned with auxiliary commonsense tasks, yields the best results in temporal commonsense inference, significantly outperforming standard fine-tuning techniques. Huang et al.~\cite{DBLP:conf/coling/Huang0HTX24} presented an approach to enhancing event causality identification, which distills commonsense meta-graph from ConceptNet and then aggregates heterogeneous information from the external meta-graph and the input text through a commonsense-aware memory network. In the end, they adopted continual pre-training and fine-tuning to further fuse the input information, and make final predictions. Wang et al.~\cite{DBLP:journals/corr/abs-2407-18479} proposed a framework called SinLG, which integrates a pre-trained language model (PLM) with a graph neural network (GNN) to enhance multi-turn response selection in dialogue systems. The model leverages PLMs for capturing word correlations and uses a GNN to incorporate commonsense knowledge from external knowledge graphs. By fusing the representations from the PLM and GNN through a Siamese network architecture~\cite{DBLP:conf/cvpr/ChenH21} and optimizing a similarity loss function, the model transfers commonsense knowledge to the PLM. This approach improves both the performance and efficiency of the model, as only the PLM is required during inference.

\subsection{Local Modification}
Local parameter editing represents an efficient and interpretable method for knowledge editing, making it possible to update open-source Transformer-based models without re-training. As shown in Fig.~\ref{fig:methodologies_local}, it aims to identify and update the specific parameters in LLMs related to targeted knowledge, allowing for the incorporation of new, relevant information. Previous applications of this technique have primarily focused on editing factual knowledge with singular correct answers. Gupta et al.~\cite{DBLP:conf/emnlp/GuptaMS00WT23} proposed MEMIT$_{\text{CSK}}$, which marks a pioneering effort in extending this approach to the editing of commonsense knowledge, typically accommodating multiple correct answers. They initiated their study with a causal analysis of GPT-2 Large and XL models~\cite{radford2019language}, establishing clear causal correlations between model predictions and local parameters at the subject, verb, and object positions when the models conduct commonsense plausibility tasks. Subsequently, they proposed Mass Editing Memory in a Transformer for Commonsense Knowledge (MEMIT$_{\text{CSK}}$). This extension encompasses editing diverse token positions (subject, verb, object) and refining the layer selection strategy. Compared to the vanilla MEMIT method~\cite{DBLP:conf/iclr/MengSABB23}, MEMIT$_{\text{CSK}}$ improves it for the commonsense domain by varying edit tokens and improving the layer selection strategy.

\section{Applications}
The application of commonsense knowledge in NLP has opened new avenues for enhancing the performance and versatility of various NLP tasks. This section explores how incorporating commonsense knowledge enables more nuanced and contextually aware language understanding, making it possible to tackle complex tasks that require a deeper level of reasoning and interpretation. We will discuss several key applications where commonsense knowledge plays a pivotal role, including emotion detection, sarcasm detection and generation, dialogue generation, question answering, machine translation, and trustworthy AI. By examining these applications, we aim to demonstrate the tangible benefits and the transformative potential of integrating commonsense reasoning into NLP models, paving the way for more sophisticated and human-like interactions with machines.

\subsection{Emotion Detection}
Commonsense knowledge, particularly intuitive psychology, enhances NLP models by providing a nuanced understanding of social and cultural contexts. This capability allows for accurate inference and prediction of contextual situations, evolving emotional states, and intentions. As a result, it significantly improves the models' proficiency in detecting subtle and fine-grained emotions. Zhong et al.~\cite{DBLP:conf/emnlp/ZhongWM19} represented a pioneering effort in integrating external commonsense knowledge for emotion detection within textual dialogues. They proposed a Knowledge Enriched Transformer (KET), where contextual utterances are interpreted using hierarchical self-attention, and external commonsense knowledge is dynamically leveraged using a context-aware affective graph attention mechanism. Based on that, Suresh and Ong~\cite{DBLP:conf/acii/SureshO21} implemented more fine-grained emotion recognition, without re-training language models from scratch by augmenting the word-embedding. Ghosal et al.~\cite{DBLP:conf/emnlp/GhosalMGMP20} introduced the COSMIC framework, designed to enhance utterance-level emotion recognition in conversations by incorporating commonsense knowledge. This knowledge encompasses aspects like mental states, events, and causal relationships, which are extracted from an external commonsense knowledge graph. Nie et al.~\cite{nie2023long} employed the classic commonsense knowledge atlas \textsc{Atomic}~\cite{DBLP:conf/aaai/SapBABLRRSC19} to retrieve prior knowledge for each utterance. This prior knowledge was subsequently integrated as auxiliary information into an growing graph model, facilitating the enhancement of utterance embeddings. Finally, they utilized a cross-attention module to amalgamate previously extracted utterance features with latent conversation topic information, which was derived through a novel self-supervised learning approach.

\subsection{Sarcasm Detection and Generation}
Sarcasm detection and generation are challenging tasks in NLP due to the subtle and often context-dependent nature of sarcastic expressions. Sarcasm frequently involves saying the opposite of what is meant, making it difficult for models to accurately interpret or generate such language without a deep understanding of context, tone, and intent. Commonsense knowledge can play a crucial role in this process by providing the model with the necessary background information to recognize when a statement is intended sarcastically and to generate sarcastic responses appropriately.

Basu Roy Chowdhury and Chaturvedi~\cite{basu-roy-chowdhury-chaturvedi-2021-commonsense} explored the use of commonsense knowledge in sarcasm detection, through a graph convolutional network~\cite{DBLP:conf/iclr/KipfW17} integrating pre-trained language model embeddings and COMET~\cite{DBLP:conf/acl/BosselutRSMCC19} (a GPT-2~\cite{radford2019language} model fine-tuned on \textsc{Atomic}~\cite{DBLP:conf/aaai/SapBABLRRSC19}) commonsense sequences. Despite applying it to multiple datasets, the commonsense-augmented model performs similarly to a baseline model. Through analysis, they found commonsense helps with sarcasm involving polarity contrast but adds little value in non-sarcastic contexts. Li et al.~\cite{DBLP:journals/taslp/LiPLFW21} presented a sarcasm detection model that integrates commonsense knowledge via COMET. Combining a BERT-based encoder with a commonsense reasoning module, the model improves sarcasm detection on Twitter and Reddit datasets. Chen et al.~\cite{DBLP:conf/smc/ChenLLZL22} introduced a sarcasm detection model using a heterogeneous graph attention network that incorporates commonsense knowledge from ConceptNet~\cite{DBLP:conf/aaai/SpeerCH17}. This approach captures implied sentiments often missed by sentiment analysis alone, outperforming previous methods on Reddit and Internet Argument Corpus datasets. Yu et al.~\cite{DBLP:conf/ijcai/Yu0WLW023} presented a sarcasm detection model, Commonsense Sentiment Dependency Graph Convolutional Network (CSDGCN), which integrates commonsense knowledge with syntactic structure. By constructing commonsense-augmented sentiment and dependency graphs, the model captures contradictions more effectively. Experiments show CSDGCN outperforms previous methods, highlighting the value of combining commonsense and syntax for sarcasm detection.

Compared to sarcasm detection, fewer studies have focused on sarcasm generation, as it is generally considered more challenging and complex. Mishra et al.~\cite{DBLP:conf/emnlp/MishraTS19} presented a framework for generating sarcastic sentences from literal, negative opinions without the need for paired training data. The system leverages a modular approach, consisting of four stages: sentiment neutralization, positive sentiment induction, negative situation retrieval, and sarcasm synthesis. Chakrabarty et al.~\cite{chakrabarty-etal-2020-r} presented an unsupervised sarcasm generation framework with three modules: valence reversal, commonsense retrieval, and semantic incongruity ranking. The model enhances non-sarcastic inputs with commonsense knowledge to generate creative, humorous sarcastic responses. Human evaluations show it outperforms previous methods, making it useful for conversational agents and content creation.

\subsection{Dialogue Generation}
Open-domain dialogue generation involves creating conversational agents capable of engaging in discussions across a wide range of topics, without being limited to a specific domain or subject area. For insights into how commonsense knowledge can enhance open-domain dialogue generation, we refer readers to Section~\ref{subsec:external_memorization}, where we discuss recent works in this area. Additionally, LLMs like GPT can be considered open-domain dialogue agents due to their ability to converse on diverse topics. Given the extensive research on LLMs, we recommend readers consult survey papers focused on general knowledge processing in these models~\cite{DBLP:journals/tkde/PanLWCWW24,DBLP:journals/corr/abs-2310-16218,DBLP:journals/corr/abs-2310-07521}. Here, we focus on a specific sub-domain of dialogue generation---empathetic dialogue generation---where commonsense knowledge plays a particularly crucial role.

Empathetic dialogue generation is a critical area of NLP that focuses on creating responses that reflect understanding and emotional sensitivity to the user's situation or feelings. Incorporating commonsense knowledge, particularly intuitive psychology---the understanding of others' emotions, intentions, and mental states---plays a vital role in enhancing the empathy of dialogue systems. By leveraging such knowledge, models can better grasp the underlying emotional context of a conversation and generate responses that are not only contextually appropriate but also emotionally resonant. Zhong et al.~\cite{DBLP:conf/aaai/ZhongWLZWM21} proposed a model that integrates both rationality, in the form of commonsense, and emotion into conversational AI systems. The model constructs and incorporates latent concepts derived from an emotion-aware commonsense knowledge graph to generate more accurate and emotionally appropriate responses. Tu et al.~\cite{DBLP:conf/acl/TuLC0W022} introduced the MISC model that integrates fine-grained emotion understanding with a mixed-strategy response model. The model leverages COMET to enhance the understanding of a user's mental state and employs a strategy codebook for generating empathetic and contextually appropriate responses. Sabour et al.~\cite{DBLP:conf/aaai/SabourZH22} also leveraged the COMET framework to produce external commonsense knowledge (inferences about the user's situation and emotional state), and proposed the CEM approach that incorporates both affective and cognitive aspects of empathy, leading to more contextually appropriate and empathetic responses. Li et al.~\cite{DBLP:conf/aaai/LiLRRC22} proposed the KEMP model, which enhances empathetic dialogue generation by constructing an emotional context graph from external knowledge (ConceptNet and NRC VAD~\cite{DBLP:conf/acl/Mohammad18}) and employing an emotional cross-attention mechanism to learn and utilize emotional dependencies for generating more empathetic responses. Cai et al.~\cite{DBLP:conf/acl/CaiSXSWGZX23} proposed a method for empathetic dialogue generation that dynamically selects and integrates the most relevant commonsense knowledge using an adaptive module, enhancing response consistency and emotional alignment with the speaker's context. Zhou et al.~\cite{DBLP:conf/acl/ZhouZW0H23} introduced the CASE model, which enhances empathetic response generation by aligning coarse-to-fine cognition and affection through the integration of a commonsense cognition graph and an emotional concept graph.

\subsection{Textual Question Answering}
Textual question answering aims at providing a precise answer to the given question in natural language, by mining from unstructured textual data. Depending on whether an accompanied block of text is provided as context, textual question answering is further divided into two sub-tasks: machine reading comprehension (MRC) and open question answering (OpenQA). In MRC, having access to external commonsense knowledge allows inference over information out of the context, even more so for OpenQA. Mihaylov and Frank~\cite{DBLP:conf/acl/FrankM18} developed a cloze-style reading comprehension model that encodes relevant commonsense knowledge from ConceptNet as a key-value memory. Other than performance improvement over the baseline, the model could also provide evidence about the background knowledge. Bauer et al.~\cite{DBLP:conf/emnlp/BauerWB18} extracted relevant multi-hop knowledge reasoning paths from ConceptNet that were subsequently used in the reasoning cells to select helpful information for the NarrativeQA~\cite{DBLP:journals/tacl/KociskySBDHMG18} task. Zhong et al.~\cite{DBLP:conf/nlpcc/ZhongTDZWY19} proposed to pre-train a commonsense scoring function that exploits the graph structures in ConceptNet to model direct and indirect relations between concepts, which can be added to existing neural-based QA models. Similarly, Chen et al.~\cite{DBLP:conf/coling/ChenJCZ20} adopted a graph-based module to iteratively retrieve concepts and entities from multiple knowledge sources, including ConceptNet. Bian et al.~\cite{DBLP:conf/aaai/BianH0021} converted the commonsense QA task into an MRC task by transforming the commonsense knowledge into a block of text to serve as the accompanied context for the model to infer the answer.

\subsection{Machine Translation}
It is believed that Bar-Hillel~\cite{DBLP:journals/ac/Bar-Hillel60} was the first to mention the importance of incorporating extra-linguistic knowledge into machine translation programs, for the purpose of resolving semantical ambiguities. Since then, many efforts have been put into the task of knowledge-based machine translation~\cite{DBLP:journals/mt/Nirenburg89}. With the advent of data-driven approaches and the creation of commonsense knowledge resources, there has been some work dedicated to the incorporation of commonsense knowledge into machine translation systems. Caseli et al.~\cite{DBLP:conf/naacl/CaseliSS10} worked on a project that aims at applying commonsense knowledge extracted from bilingual ConceptNets to the machine translation task to generate more culturally contextualized translations. To evaluate the commonsense reasoning capability of neural machine translation, He et al.~\cite{DBLP:conf/emnlp/HeWXL20} created a test suite of 1,200 triplets covering lexical and syntactic ambiguities that require commonsense reasoning to resolve. Liu et al.~\cite{DBLP:conf/acl/0002WWZYZ23} improved this evaluation method by considering the candidates as well as the commonsense entities in the candidates, making the results more aligned with human judgment. They also showed that the incorporation of pre-trained knowledge leads to better commonsense reasoning abilities.

\subsection{Trustworthy AI}
Trustworthy AI is an approach to AI development that emphasizes safety and transparency for users. In the field of NLP, one way to promote more trustworthy AI is by mitigating \emph{hallucinations} in large language models (LLMs). Hallucination in LLMs refers to the generation of text that is factually incorrect, nonsensical, or irrelevant to the given context. This occurs when a model produces content that, while seemingly coherent, is not grounded in reality or does not align with the input data. Hallucinations can significantly undermine the reliability and trustworthiness of language models, particularly in critical applications such as healthcare and finance, where LLMs like GPT-4~\cite{DBLP:journals/corr/abs-2303-08774} are increasingly integral. Consequently, mitigating hallucination is essential. In anti-hallucination research, commonsense knowledge shares characteristics with domain-specific knowledge. Due to this similarity, a common approach to reducing hallucinations in LLMs is to use general knowledge enhancement techniques without strictly differentiating between domain-specific and commonsense knowledge. This method’s strength lies in its universality and flexibility, allowing it to address various types of knowledge simultaneously.
For methods using general knowledge enhancement techniques to reduce hallucinations in LLMs, we recommend referring to the survey paper by Tonmoy et al.~\cite{DBLP:journals/corr/abs-2401-01313}. However, considering the unique role of commonsense knowledge, this subsection will focus on papers specifically addressing commonsense knowledge.

Toroghi et al.~\cite{DBLP:journals/corr/abs-2403-01390} proposed \emph{Right for Right Reasons ($R^3$)}, which leverages the intrinsic commonsense knowledge in LLMs with external knowledge graphs to mitigate hallucination. It begins by identifying query anchor entities in the input and extracting related sub-graphs. It then elicits a commonsense axiom from LLMs to guide reasoning within each search tree branch. The algorithm iteratively evaluates axiom satisfiability using available knowledge graph facts at each tree level, making every reasoning step verifiable. Similar to $R^3$, the method proposed by Li et al.~\cite{DBLP:journals/corr/abs-2402-04978} leverages external KGs and the LLMs' intrinsic knowledge. However, this approach first employs entity-guided knowledge sub-graph retrieval, and then uses a probabilistic model for step-by-step inference. Zhang et al.~\cite{DBLP:journals/corr/abs-2402-09267} proposed the \textit{Self-Alignment for Factuality} approach, leveraging LLMs' internal knowledge to generate self-knowledge-guided training data for fine-tuning, thereby enhancing commonsense reasoning and mitigating hallucinations. This method comprises three key steps: (i) sampling and verifying candidate answers for factual correctness; (ii) constructing true/false training examples encompassing both correct and incorrect responses while preserving the model's prediction distribution across diverse scenarios; and (iii) implementing Direct Preference Optimization (DPO)~\cite{DBLP:conf/nips/RafailovSMMEF23} for fine-tuning.

It is believed that maintaining contextual semantic consistency can effectively reduce hallucinations in LLMs~\cite{DBLP:journals/nature/FarquharKKG24}. Gao et al.~\cite{DBLP:journals/corr/abs-2402-17011} leveraged diffusion to learn to reconstruct the implicit semantic connections between narrative contexts and relevant commonsense knowledge by training Diffusion Commonsense Transformer models. This approach avoids the language models’ reasoning process from simply retrieving discrete factual information from a commonsense knowledge base. Instead, it converts the retrieved and contextually implied commonsense knowledge into continuous vector representations, dynamically predicting and incorporating them into the language models.

\section{Discussion}
In this discussion section, we delve into the unique contributions of this survey in comparison to existing literature, highlighting how our work differs in its comprehensive coverage of commonsense knowledge in NLP. We will also explore the challenges that remain unaddressed in the field, and propose potential future research directions to advance the integration of commonsense knowledge in NLP models. This discussion aims to position our survey within the broader landscape of research and provide insights for further developments in the domain.

\subsection{Comparison to Existing Literature}
Davis and Marcus~\cite{DBLP:journals/cacm/DavisM15} explored the challenges and importance of commonsense reasoning and knowledge in artificial intelligence, highlighting that despite advances in AI, progress in this area has been slow. They discussed various approaches, including logical analysis, handcrafted knowledge bases, web mining, and crowdsourcing, but noted that no single method has emerged as a comprehensive solution. Tandon et al.~\cite{DBLP:journals/sigmod/TandonVM17} provided a comprehensive overview of the role of commonsense knowledge in enhancing machine intelligence. They discussed the challenges in acquiring, representing, and applying commonsense knowledge in various AI tasks, highlighting its importance in NLP and smart computing applications. These two works represent early surveys of commonsense knowledge from a broader AI perspective, offering a general review of commonsense knowledge in relation to machine intelligence. In contrast, our work focuses specifically on the field of NLP, providing a comprehensive introduction to commonsense knowledge within this context and exploring its applications in NLP models, including language models.

Some surveys have focused on specific aspects of commonsense knowledge. For example, Davis~\cite{DBLP:journals/csur/Davis24} provided a comprehensive overview of over 100 benchmarks developed to assess commonsense reasoning in AI systems, highlighting significant flaws in many existing benchmarks. In the context of NLP, some surveys have focused on certain tasks or applications. For example, Storks et al.~\cite{DBLP:journals/corr/abs-1904-01172} provided a comprehensive overview of recent developments in natural language inference (NLI), focusing on the creation of benchmark datasets, the development of knowledge resources, and the emergence of state-of-the-art learning and inference models. They highlighted the critical role of these benchmarks in advancing NLI research, addressing the complexities of reasoning beyond the explicit content of text by incorporating commonsense and world knowledge. Bhargava and Ng~\cite{DBLP:conf/aaai/Bhargava022} provided a survey of the use of pre-trained language models (PLMs) for commonsense knowledge reasoning and generation, and discussed the strengths and limitations of current PLMs in various tasks. Richardson and Heck~\cite{DBLP:journals/corr/abs-2302-07926} provided a comprehensive survey on the integration of commonsense reasoning in conversational AI, and Liu et al.~\cite{DBLP:conf/acl/0002WWZYZ23} presented a study on the integration of commonsense reasoning in neural machine translation. While these surveys have different focuses, our paper is more comprehensive, providing a holistic and up-to-date overview that integrates various aspects of NLP commonsense reasoning, including resources, benchmarks, methodologies, and applications.

\subsection{Future Research Directions}
Commonsense reasoning continues to pose a major challenge in NLP due to the vast and varied nature of human commonsense knowledge, which makes complete integration difficult. Furthermore, commonsense knowledge differs significantly across contexts, adding to the complexity. While manually constructing a commonsense knowledge base or using crowdsourcing is time-consuming and expensive, automatic extraction methods often face issues with accuracy. Despite these obstacles, there are promising future directions in commonsense reasoning research that hold the potential to provide valuable insights for the NLP community.

\subsubsection{Expression of commonsense knowledge}
As shown in Table~\ref{tab:knowledge_bases}, early works on commonsense knowledge bases (prior to 2020) typically adopted a structured graph format, where concepts are represented as nodes and their relationships as edges, such as in ConceptNet~\cite{DBLP:conf/aaai/SpeerCH17} and \textsc{Atomic}~\cite{DBLP:conf/aaai/SapBABLRRSC19}. More recent works, like MickeyCorpus~\cite{DBLP:conf/acl/LinLQ020} and Moral Events~\cite{zhang-etal-2024-moka}, express commonsense knowledge directly in natural language, i.e., unstructured text, often with annotations. This shift is likely due to the advent and development of LLMs, as the most effective way to process commonsense knowledge with LLMs is through natural language interaction. In fact, a study by Bian et al.~\cite{DBLP:conf/acl/BianHL0H024} demonstrated that LLMs retrieve commonsense knowledge more confidently and accurately from unstructured stories than from structured rules. Additionally, LLMs are more effective at leveraging stories than rules for commonsense reasoning. Therefore, further investigation into the optimal ways of expressing and utilizing different types of commonsense knowledge for LLMs is warranted.

\subsubsection{Knowledge priorities and conflicts in PLMs}
The emergence of pre-trained large language models, such as ChatGPT, has expanded the use cases of language models, resulting in situations where models must handle various types of knowledge (commonsense, domain-specific, and external contextual knowledge) within the same reasoning task. This convergence of knowledge types can potentially result in knowledge conflicts, ultimately causing hallucinations during the model's reasoning process. While researchers like Jin et al.\cite{DBLP:conf/acl/JinCY0XLJ0024}, Wu et al.\cite{DBLP:conf/acl/Wu0CWRKRM24} and Zhang et al.\cite{DBLP:journals/corr/abs-2405-19010} have astutely identified and discussed this phenomenon, current literature primarily focuses on detecting and resolving conflicts between the model's internal parametric knowledge and user-provided external non-parametric knowledge. We advocate for a deeper investigation into knowledge prioritization within language models, as well as the detection and resolution of internal parametric knowledge conflicts that arise after augmenting the model with various types of knowledge. This research direction could lead to more coherent and reliable reasoning in LLMs, potentially mitigating hallucinations and improving performance across diverse tasks that require the integration of multiple knowledge types.

\subsubsection{Commonsense reasoning benchmarks}
As LLMs have advanced significantly in recent years, numerous benchmarks have been developed to assess their various capabilities. However, a study by McIntosh et al.~\cite{DBLP:journals/corr/abs-2402-09880} analyzed 23 state-of-the-art benchmarks and highlighted significant limitations, including biases, challenges in accurately measuring reasoning abilities, and a lack of attention to cultural and ideological diversity. Bhargava and Ng~\cite{DBLP:conf/aaai/Bhargava022} further noted that many LLMs have achieved near-human performance on commonsense reasoning benchmarks, particularly in question answering tasks. However, high scores on these benchmarks do not necessarily indicate that the models possess near-human commonsense reasoning abilities. A deeper analysis of models' true reasoning capabilities is needed, along with the development of more comprehensive and standardized benchmarks to keep pace with the evolving nature of LLMs.

\subsubsection{Commonsense knowledge in multicultural contexts}
While naive physics commonsense knowledge is generally universal across human societies, intuitive psychology commonsense can vary depending on linguistic or cultural backgrounds, particularly regarding daily activities, social behaviors, and norms. Some existing studies have incorporated multilingual settings when developing commonsense knowledge resources and benchmarks. For instance, XCOPA~\cite{DBLP:conf/emnlp/PontiGMLVK20} is a multilingual benchmark that translates and re-annotates the English COPA~\cite{DBLP:conf/aaaiss/RoemmeleBG11} into 11 languages. Similarly, X-CSQA and X-CODAH~\cite{DBLP:conf/acl/LinLQ020} are translations of the English CSQA~\cite{DBLP:conf/naacl/TalmorHLB19} and CODAH~\cite{DBLP:journals/corr/abs-1904-04365} benchmarks, with questions that might involve cultural differences intentionally removed. However, simply translating existing resources is insufficient for capturing the nuances of multilingual commonsense knowledge, as noted by Sakai et al.~\cite{DBLP:conf/acl/0010KW24}. Instead of relying on translation, they built on a multilingual commonsense knowledge base (ConceptNet) and used LLMs to generate and validate questions, options, and additional distractors. Despite these efforts, the differences in social behaviors and norms across multicultural contexts remain underexplored. Datasets like \textsc{SocialDial}~\cite{DBLP:conf/sigir/ZhanLWLFKHQSSZS23}, which focuses on Chinese social culture, and \textsc{NormDial}~\cite{DBLP:conf/emnlp/LiSSCM23}, which studies social norms in both Chinese and American cultures in conversational settings, are important steps forward. Expanding commonsense knowledge resources and benchmarks to encompass more diverse cultural backgrounds is crucial, and this task will likely require collaboration with experts in sociology and linguistics.

\subsubsection{Commonsense knowledge in multimodal models}
Recent advancements in multimodal models have significantly improved the integration of commonsense knowledge across various modalities. Research on Commonsense-T2I~\cite{DBLP:journals/corr/abs-2406-07546} demonstrates that enhancing commonsense knowledge in a single modality is insufficient to ensure robust commonsense reasoning capabilities in large multimodal models. Therefore, enhancing, aligning and resolving conflicts between different modalities remains challenging. We call upon the research community to further investigate commonsense enhancement methods for multimodal models, particularly focusing on aligning commonsense knowledge across different modalities to prevent internal knowledge conflicts. While there have been initial forays into addressing this challenge, including the proposal of a universal representation framework by Zhang et al.~\cite{DBLP:journals/pami/ZhangCWUSLZ23}, the field is still in the nascent stages of understanding how commonsense knowledge can be effectively represented and utilized within multimodal LLMs. Additionally, due to modern (multimodal) language models' deployment constraints on resource-limited devices such as smartphones, we advocate for distilling commonsense-enhanced multimodal large language models into smaller multimodal or unimodal models. We encourage the community to continue investigating techniques to improve inference abilities in resource-limited deployed models.

\section{Conclusion}
In conclusion, this survey has provided a comprehensive overview of the integration of commonsense knowledge in natural language processing, covering key aspects such as prominent knowledge bases, benchmarks for assessing commonsense reasoning, methodologies for incorporating this knowledge into models, and real-world applications. By synthesizing the current state of research, we have highlighted the advancements and identified the gaps that still exist in the field. As NLP continues to evolve, the role of commonsense knowledge will become increasingly crucial, offering opportunities for more intelligent and context-aware systems. Future research should focus on addressing the identified challenges, particularly in enhancing the scalability and generalization of models, to further bridge the gap between human-like understanding and machine learning.

\bibliographystyle{IEEEtran}
\bibliography{references}

\end{document}